\documentclass[conference]{IEEEtran}
\IEEEoverridecommandlockouts
\usepackage{amsmath,amssymb,amsfonts}
\usepackage{algorithmic}
\usepackage{graphicx}
\usepackage{textcomp}
\usepackage{xcolor}
\usepackage{caption}
\usepackage{setspace, lipsum}
\usepackage{hyperref}
\usepackage{multirow}
\usepackage{color, colortbl}
\usepackage{url}
\usepackage{subcaption}
\usepackage{float}
\usepackage{nccmath}
\usepackage{amsthm,bbm}
\usepackage{adjustbox}
\usepackage{bbding, pifont}
\hypersetup{
    colorlinks,
    linkcolor={red!50!black},
    citecolor={blue!50!black},
    urlcolor={blue!80!black}
}

\definecolor{darkred}{rgb}{0.7, 0, 0}
\definecolor{darkblue}{rgb}{0.0, 0.0, 0.7}
\definecolor{darkgreen}{rgb}{0, 0.4, 0}
\newcommand{\cmark}{\textcolor{darkgreen}{\ding{51}}}
\newcommand{\xmark}{\textcolor{darkred}{\ding{55}}}
\newcommand{\best}[1]{\color{red}\textbf{#1}}
\newcommand{\second}[1]{\color{blue}\textbf{#1}}

\newcommand{\Gen}{\mathcal{G}}
\newcommand{\Dis}{\mathcal{D}}
\newcommand{\Cl}{\mathcal{C}}

\newcommand{\ms}{m^S}
\newcommand{\mr}{m^R}
\newcommand{\xs}{x^S}
\newcommand{\xr}{x^R}

\newcommand{\SynData}{\mathbf{D}^S}
\newcommand{\RealData}{\mathbf{D}^{R}}
\usepackage[sorting = none, backend = bibtex, style=numeric-comp]{biblatex}
\AtBeginBibliography{\small}

\begin{document}

\title{Towards Pragmatic Semantic Image Synthesis for Urban Scenes

\thanks{\textsuperscript{*} Bot authors equally contributed to thus work\\ Corresponding author email: george.eskandar@iss.uni-stuttgart.de}
\thanks{The research leading to these results is funded by the German Federal Ministry for Economic Affairs and Energy within the project "AI Delta Learning." The authors would like to thank the consortium for the successful cooperation.}
\thanks{Code is available at \url{https://github.com/GeorgeEskandar/Towards-Pragmatic-Semantic-Image-Synthesis-for-Urban-Scenes}}
}

\author{\IEEEauthorblockN{George Eskandar\textsuperscript{1}\textsuperscript{*}, Diandian Guo\textsuperscript{*}\textsuperscript{1}, Karim Guirguis\textsuperscript{2}, Bin Yang\textsuperscript{1}}
\IEEEauthorblockA{\textsuperscript{1}Institute of Signal Processing and System Theory, University of Stuttgart, Stuttgart, Germany} 
\IEEEauthorblockA{\textsuperscript{2}Bosch Center for Artificial Intelligence, Renningen, Germany} 
}
\maketitle
\begin{abstract}
The need for large  amounts of training and validation data is a huge concern in scaling AI algorithms for autonomous driving. Semantic Image Synthesis (SIS), or label-to-image translation, promises to address this issue by translating semantic layouts to images, providing a controllable generation of photorealistic data. However, they require a large amount of paired data, incurring extra costs. In this work, we present a new task: given a dataset with synthetic images and labels and a dataset with unlabeled real images, our goal is to learn a model that can generate images with the content of the input mask and the appearance of real images. This new task reframes the well-known unsupervised SIS task in a more practical setting, where we leverage cheaply available synthetic data from a driving simulator to learn how to generate photorealistic images of urban scenes. This stands in contrast to previous works, which assume that labels and images come from the same domain but are unpaired during training. We find that previous unsupervised works underperform on this task, as they do not handle distribution shifts between two different domains. To bypass these problems, we propose a novel framework with two main contributions. First, we leverage the synthetic image as a guide to the content of the generated image by penalizing the difference between their high-level features on a patch level. Second, in contrast to previous works which employ one discriminator that overfits the target domain semantic distribution, we employ a discriminator for the whole image and multiscale discriminators on the image patches. Extensive comparisons on the benchmarks GTA-V $\rightarrow$ Cityscapes and GTA-V $\rightarrow$ Mapillary show the superior performance of the proposed model against state-of-the-art on this task.     

\end{abstract}

\begin{IEEEkeywords}
GANs, Image-to-Image Translation, Semantic Image Synthesis, Synthetic Data, Unsupervised Learning
\end{IEEEkeywords}

\section{Introduction}
\label{sec:intro}
Semantic Image Synthesis is a subclass of Image-to-Image translation (I2I) where an image is generated from a semantic layout. In the context of autonomous driving, SIS is a promising method for generating diverse training and validation data because it can provide photorealism and controllability, two essential qualities for data augmentation schemes. Photorealism means that the generated images have the same texture and appearance as images recorded in real life; otherwise, a domain gap will ensue from the difference between training and test time distributions. On the other hand, controllability means that the user can generate or edit an image with predefined characteristics (changing weather, adding or removing cars, changing road width, increasing the number of pedestrians ...). Having user control over the input RGB space is of special interest for model validation~\cite{explainability} because it helps understand an AI model's decision through counterfactual reasoning: for instance, we wish to know whether a perception algorithm would behave correctly had there been more cars or pedestrians in a given scene. In this regard, SIS allows easy and explicit user control of scene editing through the manipulation of the input semantic map. For instance, one can add, delete or alter the style of different objects. SIS has been pivotal in several frameworks~\cite{palette, sbgan, decomposing} where semantic layouts are used as a prior for urban scene generation. 

\begin{figure}[t]
\centering
\begin{center}
\includegraphics[width=0.95\linewidth]{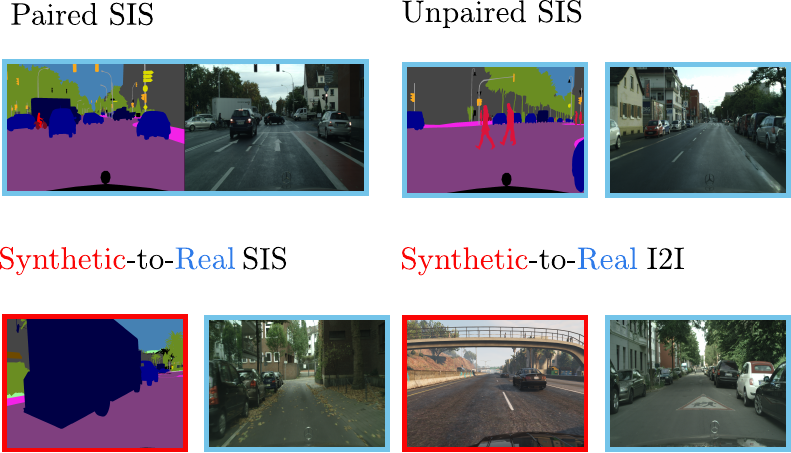}
\end{center}
\caption{{\small Overview of different SIS and I2I paradigms: (a) Paired SIS, (b) Unpaired SIS: the common setting in the previous works was to assume images and labels come from the same distribution but they are unpaired, (c) Synthetic-to-Real SIS: the proposed task, labels come from a synthetic data, images originate from a real dataset, (d) Synthetic-to-Real I2I: source and target domains are different, but images are used in both domains, which is considerably easier than using labels as input, but does not provide controllability.} }
\label{fig:teaser}
\end{figure}
\begin{table}[t!]
\setlength{\tabcolsep}{0.2em}
\centering

\adjustbox{width=0.99\linewidth}{\begin{tabular}{|c|c|c|c|c|}
    \hline
    Task & Photorealism & Controllability & Cross-Domain & Cheap data \\
    \hline
    Syn-to-Real I2I & \cmark & \xmark & \cmark & \cmark \\
    Paired SIS  & \cmark & \cmark & \xmark & \xmark \\
    Unpaired SIS                        & \cmark & \cmark & \xmark & \xmark \\
    Syn-to-Real SIS               & \cmark & \cmark & \cmark & \cmark \\
    \hline
\end{tabular}}
\caption{{\small Comparison between different SIS and I2I paradigms: the proposed task combines several practical advantages compared to the previous tasks.}}
\label{table:teaser}
    \vspace{-1.5em}
\end{table}

However, the necessity for a significant amount of paired training data undermines the initial intent of SIS, which was to provide inexpensive data augmentation. Unsupervised approaches~\cite{cyclegan, munit, drit, travelgan, distancegan, gcgan, cut, srunit, sesim, src, usisicassp, usiscag} have shown impressive results without needing paired data, but because they rely on ideal assumptions, their usefulness in the actual world is limited. Mainly, they employ paired datasets (like Cityscapes) and present the images and labels to the model in an unpaired way in each training iteration mimicking a real unpaired setting. This has brought us to ask the following questions: \textit{what would a truly unpaired training setting for SIS look like? and how to pragmatically collect cheap semantic layouts for training? }

In this work, we propose a new task, Synthetic-to-Real SIS, where a mapping is learned from a set of synthetic labels to a set of real images (Figure~\ref{fig:teaser}). Since labels and images originate from 2 different domains, not only is the setting truly unpaired but also pragmatic because producing semantic layouts from a driving simulator \cite{carla}, or a graphics engine~\cite{gtav} is inexpensive (Table~\ref{table:teaser}). However, this task also introduces a new challenge: it is almost inevitable that the source and target datasets (labels and images, respectively) will have different class distributions. For instance, the synthetic dataset might contain a large proportion of buildings, while the real dataset has fewer buildings and more trees. Cross-domain differences could potentially lead to undesired semantic misalignments in the generated images (such as generating trees instead of buildings) because the model tries to imitate the target domain and ignores the source label map. This would undermine the utility of unpaired generative models.

We show that unpaired GANs underperform on the Synthetic-to-Real SIS task, as they ignore cross-domain semantic mismatches. To this end, we introduce a new framework that bypasses this limitation. The key idea of our framework is to learn to generate an image with the appearance of real images but with the content of the synthetic image that corresponds to the input label. In other words, we use the synthetic image as a \textit{guide} to the content of the generated image by leveraging high-level features of a pre-trained network. Our contributions can be summarized as follows: (1) We propose Synthetic-to-Real SIS, a new task that allows training an SIS model in a pragmatic and low-cost way, (2) we develop a new framework for this task that exploits the similarity between generated and synthetic \textit{patches} to preserve alignment with the input layout, and (3) in contrast to previous works which use one discriminator, we employ multiple discriminators on both the \textit{global} and \textit{local} image contexts to prevent overfitting on simple visual cues in the target domain. Experiments on 2 benchmarks, GTA-V $\longrightarrow$ Cityscapes and GTA-V $\longrightarrow$ Mapillary, show the superior performance of our model compared to state-of-the-art approaches.


\section{Related Works}
\label{sec:rw}

\noindent\textbf{Semantic image synthesis} is the task of translating semantic layouts to images \cite{pix2pixhd}. In contrast to I2I~\cite{pix2pix}, SIS is a severely under-constrained task because the input layout has fewer details than the output image, which is rich in high spatial frequencies like edges, corners, and texture. SPADE~\cite{spade} has made a breakthrough in this task by designing spatially adaptive normalization layers. Since then, a  plethora of frameworks~\cite{clade, sean, ccfpse, oasis, edgesis} has presented progressive architectural improvements to enhance the fidelity and alignment of generated images. 

\begin{figure}[t]
\centering
\begin{center}
\includegraphics[width=0.95\linewidth]{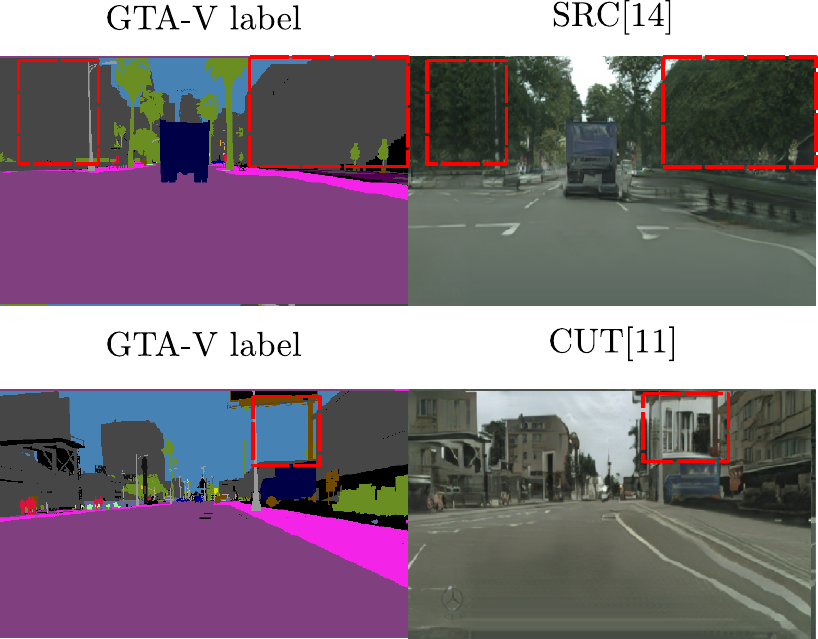}
\end{center}
\caption{{\small Performance of state-of-the-art unpaired models on Synthetic-to-Real USIS. Semantic inconsistencies (in red) can appear between the label and images.} }
\label{fig:study}
\vspace{-1em}
\end{figure}
\noindent\textbf{Unpaired image-to-image translation} is the translation of one image collection (source domain) to another (called target domain). There have been 2 main approaches to unpaired I2I: cycle consistency losses~\cite{cyclegan, munit, drit} and relationship preservation constraint~\cite{travelgan, distancegan, gcgan, cut, sesim, srunit, src}. In our previous work, we designed a framework, USIS~\cite{usiscag, usisicassp}, that achieves state-of-the-art results on the unpaired label-to-image translation task. In this work, we try to extend USIS to a more realistic unpaired scenario for urban scene generation. 

\noindent\textbf{Synthetic to real translation} is one application of unpaired I2I, where a synthetic image (produced from a simulation environment) is translated to a photorealistic image~\cite{cyclegan, epe, cut, munit}. Our approach is aligned with this line of work but uses synthetic layouts instead of images as input. This is because semantic layouts are more abstract and, thus, more manipulable than images, which opens the door to many applications such as semantic editing, model validation, and domain adaptation.  


\section{Methodology}
\label{sec:method}
\begin{figure*}[t]
\centering
\begin{center}
\includegraphics[width=0.99\textwidth]{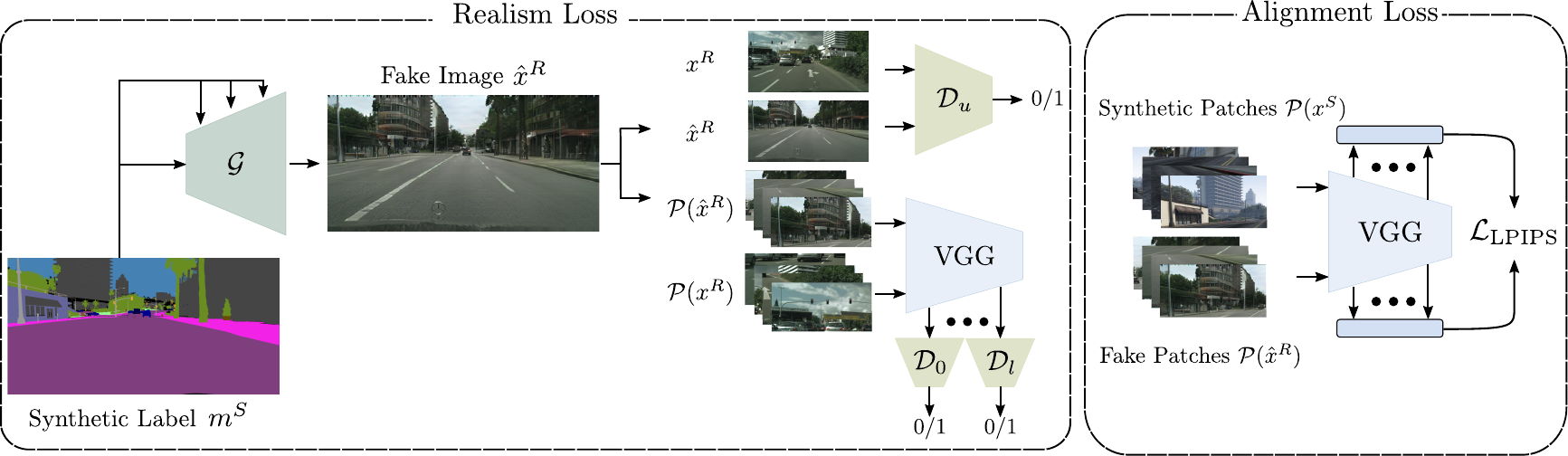}
\end{center}
\caption{An overview of the proposed unpaired framework. \textbf{Left: } We use a wavelet-based whole image discriminator and a discriminator ensemble in the high-level feature space to evaluate the realism of patches. \textbf{Right:} We use the synthetic image as a guide to promote better alignment with the semantic layout on a patch-level. }
\label{fig:framework}
\vspace{-1em}
\end{figure*}
\textbf{Problem Formulation.} In SIS, our goal is to learn a mapping from a semantic label map, $m \in \mathcal{M}$ and a noise vector $z \in \mathcal{N}(0,1)$ to a photorealistic RGB image, $\xr \in \mathcal{X}^R$. The labels $m$ are one-hot encoded with $\Cl$ classes. In Synthetic-to-Real SIS, we exploit synthetic data from graphics engines and driving simulator to obtain images $\xs$ and labels $\ms$ in a cheap and automated fashion~\cite{gtav, carla}, as the real images $\xr$ are much more costly to annotate. Thus, we are given two datasets for training: a synthetic dataset, $\SynData = \{ (\xs_i, \ms_i) \}_{i=1}^{N_S}$ and a real dataset, $\RealData = \{ \xr_j \}_{j=1}^{N_R}$. A model is trained to map a label $\ms$ to an image $\xr$. Note that, unlike I2I models, we do not use synthetic images $\xs$ as input. Conditioning the model on semantic layouts only allows maintaining easy controllability over scene generation, which is the goal of SIS~\cite{sbgan, spade}. The output of the generator can be expressed as: $\hat{x}^R = \Gen(m, z)$.

\subsection{Effect of semantic mismatch on model performance}

Since these two datasets are collected independently, there exist discrepancies in the class distribution. For instance, $\RealData$ can have more trees, while $\SynData$ can have more buildings. As the labels for real images,$\mr$, are absent during training, we do not have an \textit{a priori} knowledge of the class distribution of $\mathcal{X}^R$. To study the effect of cross-domain differences on generative models, we train 2 state-of-the-art models,  CUT~\cite{cut} and SRC~\cite{src}, on the Synthetic-to-Real task. We use GTA-V~\cite{gtav} and Cityscapes~\cite{cordts2016cityscapes} as the source and target domains. Results show that many semantic inconsistencies, such as generating trees or buildings instead of sky, appear in the generated images. Looking at the Cityscapes dataset shows that, in fact, the class 'sky' occupies a small part of the image on average, while this is not the case in GTA-V. This would suggest that previous works learn to sometimes ignore the semantic layout instead to better approximate the target domain distribution. 

\textbf{Overview of the proposed framework.} Motivated by these findings, we present a framework named to bypass cross-domain differences. The framework trains a generator $\Gen$ to satisfy two objectives: realism and alignment. Realism means that the generated images should have the appearance and texture of real images $\xr$, while the alignment objective states that the image should be aligned to the input semantic layout (see Figure~\ref{fig:framework}). In the following, we describe how we accomplish the 2 objectives for the challenging use case of synthetic to real SIS.  

\subsection{Realism Objective}
In previous works on unpaired GANs~\cite{cyclegan, cut, src, sesim, srunit}, a discriminator is used to increase the realism of generated images. An important design element in discriminators is its visual receptive field: what should the discriminator look at in the generated and real images to judge their realism? In previous works, there have been two different answers to this question: patch discrimination and whole image discrimination. The first approach outputs a matrix of realism scores, while the latter outputs only one score per image. In I2I, many works~\cite{cyclegan, cut, munit, srunit, drit, src, epe} used a patch discriminator~\cite{pix2pix} that judges the quality of individual image patches. On the other hand, it has been shown that whole image discrimination is more suitable for \textit{under-constrained} conditional GANs like SIS~\cite{usiscag}. Under-constrained means that the output image contains more details than the input layout. In this case, the discriminator needs to look at a larger spatial context to give stronger feedback to the generator.

However, both strategies become suboptimal in unpaired SIS when the distribution of $\mathcal{X}^R$ is different than $\mathcal{M}^S$. We find that the whole image discriminator of USIS focuses on a small subset of visual cues that characterizes the distribution of $\mathcal{X}^R$. For instance, the discriminator might observe that $\mathcal{X}^R$ contains a lot of trees and thus would push the generator to generate images with many trees, regardless of the input layout. This creates a semantic mismatch between the label and generated image, which is undesirable. On the other hand, we still observe that the patch discriminator does not encourage the generation of images with realistic texture. Instead, it is desirable that the discriminator learns the appearance or texture of the real images, not their semantic content, which should be determined by the input semantic map.

To overcome the shortcomings of both approaches, we draw inspiration from how a human would qualitatively judge the realism of an image. A human would first glance at the overall image and intuitively feel whether it is real or fake. Then, the human would closely inspect details of different scales, starting from more obvious and larger ones to smaller ones, to judge their quality. To realize this strategy, we first use a wavelet-based unconditional discriminator~\cite{usiscag, Gal2021SWAGANAS}, which we denote by $\Dis_u$ to evaluate the realism of the generated images. Then, we employ a discriminator ensemble, $\{\Dis_l\}_{l=1}^L$, to evaluate the realism of feature maps extracted from a pre-trained $\mathrm{VGG}$ network~\cite{vgg}, which is frozen during training. Each discriminator processes its input feature map to a one-channel tensor featuring a realism score for each pixel. We use one discriminator on each of the last $L$ ReLu layers of $\mathrm{VGG}$. Each discriminator contains 5 layers consisting of (convolution-group normalization-ReLu). We use spectral normalization on all discriminators $\Dis_l$ of the high-feature space and R1 regularization on the whole-image discriminator~\cite{stylegan}. 

Discriminating high-level features has been used in previous works~\cite{projected, epe} to focus on semantics instead of low-level details. While our work shares the same discriminator design as EPE~\cite{epe}, it differs in the regularization method. Specifically, instead of adding $1\times1$ convolution like ~\cite{projected}, or using the adaptive backpropagation of ~\cite{epe}, we regularize the discriminators by changing the input to the $\mathrm{VGG}$ extractor from the whole image to a stack of patches of the image. We define $\mathcal{P}$ as the patch operator of the image $x$, which divides it into 4 patches with equal areas and stacks them along the batch dimension of the tensor. We define $\phi_l(x) = \mathrm{VGG}_l(\mathcal{P}(x))$, as the feature of the $l$-th layer of the $\mathrm{VGG}$ network resulting from $\mathcal{P}(x)$. We argue that by providing smaller patches of the image, the $\mathrm{VGG}$ features are more fine-grained and more descriptive, because they depict a fewer number of objects than the entire image. This allows the discriminator ensemble to focus on more varied features, so it does not overfit the features of frequent and/or large classes. The adversarial learning objectives for the discriminators are: 

{\small
\begin{align}
  &\mathcal{L}_{adv_{D_l}} = -\mathbb{E}_{\xr}\left[ \log(\Dis_l(\phi_l(\xr))) \right] -\mathbb{E}_{\ms}\left[ \log(1-\Dis_l(\phi_l(\hat{x}^R))) \right], \notag\\
  &\mathcal{L}_{adv_{D_0}} = -\mathbb{E}_{\xr}\left[ \log(\Dis_u(\xr)) \right] -\mathbb{E}_{\ms}\left[ \log(1-\Dis_u(\hat{x}^R)) \right]. 
\label{eq:adv_discriminators}
\end{align}
}%
The generator's adversarial loss becomes:
{\small
 \begin{align}
\mathcal{L}_{adv_{G}} = -\mathbb{E}_{\ms}\left[ \log(\Dis_u(\hat{x}^R)) + \sum_{l=1}^{L} \log(\Dis_l(\phi_l(\hat{x}^R)))\right].
 \label{eq:adv_gen}
 \end{align}
}

We summarize our technical contributions in this part as follows:
\begin{itemize}
    \item Unlike multi-scale discriminators in previous I2I applications~\cite{pix2pixhd, ganinpainting}, which only consider patches of different resolutions, we employ a council of discriminators for the whole image and each of its patches simultaneously.
    \item Different than ~\cite{epe, projected}, we provide regularization for the high-level feature discriminators by reducing the input to the $\mathrm{VGG}$ network though the defined operator $\mathcal{P}$.
\end{itemize}

\subsection{Alignment Objective}

In SIS, it is necessary to preserve the faithfulness of the generated image to the semantic layout. In our previous work, USIS~\cite{usiscag}, this objective was achieved through a U-Net ~\cite{Ronneberger2015UNetCN} segmentation network $\mathcal{S}$ that learns to segment the generated image into its input layout. However, when a domain shift is introduced between source and target domains, the cycle segmentation loss becomes weaker than the adversarial loss, leading to deterioration in the mIoU score (see Table \ref{table:study}). 

Our key idea to remedy this issue is to rely on the synthetic image $\xs$ as a conditional guide for the generated image. Most importantly, we only wish to transfer the content of $\xs$ to $\hat{x}^R$, not its texture. To this end, we use a perceptual loss, aka LPIPS loss, between $\hat{x}^R$ and $\xs$. LPIPS has been extensively used~\cite{Gatys2016, Alexey2016} to penalize structural differences between a generated and a reference image. It computes the similarity between the $\mathrm{VGG}$ features extracted from the 2 images. 

Applying LPIPS loss between $\xs$ and $\hat{x}^R$ instead of the cycle segmentation loss of USIS leads to an improvement in the alignment score. However, we find that the alignment loss might often ignore small objects in the image. As a remedy, we propose to employ perceptual loss on a patch level instead of the global level. The alignment objective then becomes:

{
 \begin{align}
& \mathcal{L}_{\mathrm{LPIPS}} = \sum_l \left(\phi_l(\mathcal{P}(\xs)) -\phi_l(\mathcal{P}(\hat{x}^R)) \right)^2,
 \end{align}
}
where $\phi_l$ is the activation of the patches of the image from layer $l$ in the $\mathrm{VGG}$-network. Our motivation for applying a patchwise LPIPS is to amplify the alignment loss for smaller classes. A small object in an image would have a negligible contribution in the high-level representation of the whole image but would have a bigger contribution in the high-level representation of only a local part of that image.

\section{Experiments}
\label{sec:experiments}

\textbf{Datasets.} We establish 2 benchmarks using 3 datasets: Cityscapes~\cite{cordts2016cityscapes}, GTA-V~\cite{gtav} and Mapillary~\cite{mapillary}. In the first benchmark, GTA-V $\longrightarrow$ Cityscapes, we use 2 training sets: GTA-V labels and Cityscapes images. In the second benchmark, we use GTA-V labels and Mapillary images. Cityscapes contains street scenes in German cities with pixel-level annotations, while Mapillary contains more diverse street scenes around the world. GTA-V is a dataset containing $~25k$ synthetic annotated images extracted from a computer game. For all 3 datasets, we use only the 34 classes defined by Cityscapes. For both experiments, we use the same image resolution ($256 \times 512$) as USIS~\cite{usisicassp}. We use the last $5k$ images in the GTA-V dataset as a test split. We use a batchsize of 2, and a learning rate is 0.0001 in all experiments. 

\begin{table*}[t!]
\setlength{\tabcolsep}{0.2em}
\renewcommand{\arraystretch}{0.95}
\centering

\adjustbox{width=0.9\textwidth}{\begin{tabular}{|c|ccc|cc|ccc|cc|}
    \hline
    \multirow{2}{*}{Method} & \multicolumn{3}{c|}{GTA-V $\longrightarrow$ Cityscapes} &\multicolumn{2}{c|}{Cityscapes Val. Split} & \multicolumn{3}{c|}{GTA-V $\longrightarrow$ Mapillary} &  \multicolumn{2}{c|}{Mapillary Val. Split}    \\
    & \textbf{FID$\downarrow$} & \textbf{mIoU}$\uparrow$ & \textbf{KID}$\downarrow$ & \textbf{FID} $\downarrow$  &  \textbf{mIoU}$\uparrow$ & \textbf{FID}$\downarrow$ & \textbf{mIoU} $\uparrow$ & \textbf{KID}$\downarrow$ & \textbf{FID}$\downarrow$  &  \textbf{mIoU}$\uparrow$ \\
    
    \hline 

    CUT~\cite{cut}         & 66.6 & \second{21.0} & 0.029 &75.4 & \second{29.6} & 66.5 & 6.0 & 0.047 &73.4 & 16.4 \\
    SRUnit~\cite{srunit}   & 67.7 & 16.7 & 0.028 &82.1 & 23.4          & 67.4 & 9.2 & 0.042 & 70.8 & \second{16.5}\\
    F-SeSim~\cite{sesim}   & 73.3 & 15.9 & 0.031 & 111.1 & 24.4        & 133.4 & 8.9 & 0.130 & 92.9 & 15.5 \\
    SRC~\cite{src}         & 70.9 & 16.7 & 0.035 & 78.6 & 26.0         & 96.4 & 5.9 & 0.080 & 106.6 & 10.4    \\
    USIS~\cite{usisicassp} & \second{42.2} & 19.2 & \best{0.010} &\best{54.5} & 28.3 & \second{48.2} & 9.7 & \best{0.020} & \best{40.8} & 16.2 \\
    
    \hline
    Ours                   & \best{40.4} & \best{27.1} & \second{0.012} & \second{67.3} & \best{40.6} & \best{46.2} & \best{15.5} & \second{0.023} & \second{44.2} & \best{24.0} \\
    \hline

\end{tabular}}
\caption{Comparison against SOTA methods on two benchmarks.Best results are denoted in red, while second best are denoted in blue.} 
\label{table:sota}
\end{table*}

\begin{table*}[h!]
\setlength{\tabcolsep}{0.2em}
\renewcommand{\arraystretch}{0.95}
\centering

\begin{tabular}{|c|c|c|ccc|}
    \hline
    Config & Generated Output during training & Alignment method & FID $\downarrow$  &  mIoU$\uparrow$ & KID $\downarrow$  \\
    \hline 
    A & Whole Image & LPIPS on image level            & 60.5 & 21.8 & 0.0265\\
    B & Whole Image & LPIPS on patch-level (4) & \textbf{40.4} & \textbf{27.1} & \textbf{0.0120}\\
    C & Whole Image &LPIPS on patch-level (16) & 55.1 & 24.9 & 0.0181\\
    D & Individual Patches & LPIPS on patch level & 102.4 & 19.4 & 0.0641 \\
    \hline
\end{tabular}
\caption{Comparison between different alignment strategies.}
\label{table:abl_alignment}
    \vspace{-1em}
\end{table*}
\noindent \textbf{Metrics.} A good generative model should generalize well on both synthetic and real labels, meaning it should perform well, whether the input to the model is a synthetic map $\ms$, or a real map, $\mr$. We use the Frechet Inception Distance (FID) and Kernel Inception Distance (KID) to measure image fidelity and mean Intersection over Union to measure the alignment between the generated images and the input labels. For mIoU calculation, a DRN-D-105~\cite{yu2017dilated} pre-trained on the corresponding real dataset is used. We perform 2 sets of experiments: in the first, we train using GTA-V labels and Cityscapes images, and in the second, we use GTA-V labels and Mapillary images. To test the performance of each set of experiments, we use two test splits: 1) the first test split consists of the labels and images in the official validation splits of the corresponding real dataset (Cityscapes/Mapillary), 2) the second test split consists of GTA-V labels in the test split (last $5k$ labels in the GTA-V dataset) and all images in the validation split of the real dataset. mIoU is always computed between the generated images and input labels, while FID is computed between the generated images and the images in the test split. In the tables, we denote test split 1 by the corresponding real dataset's name and test split 2 by GTA-V $\rightarrow$ Cityscapes/Mapillary.

\section{Results}
\label{sec:results}
\subsection{Comparison against state-of-the-art} 
\begin{figure*}[h!]
\centering
\includegraphics[width=0.95\textwidth]{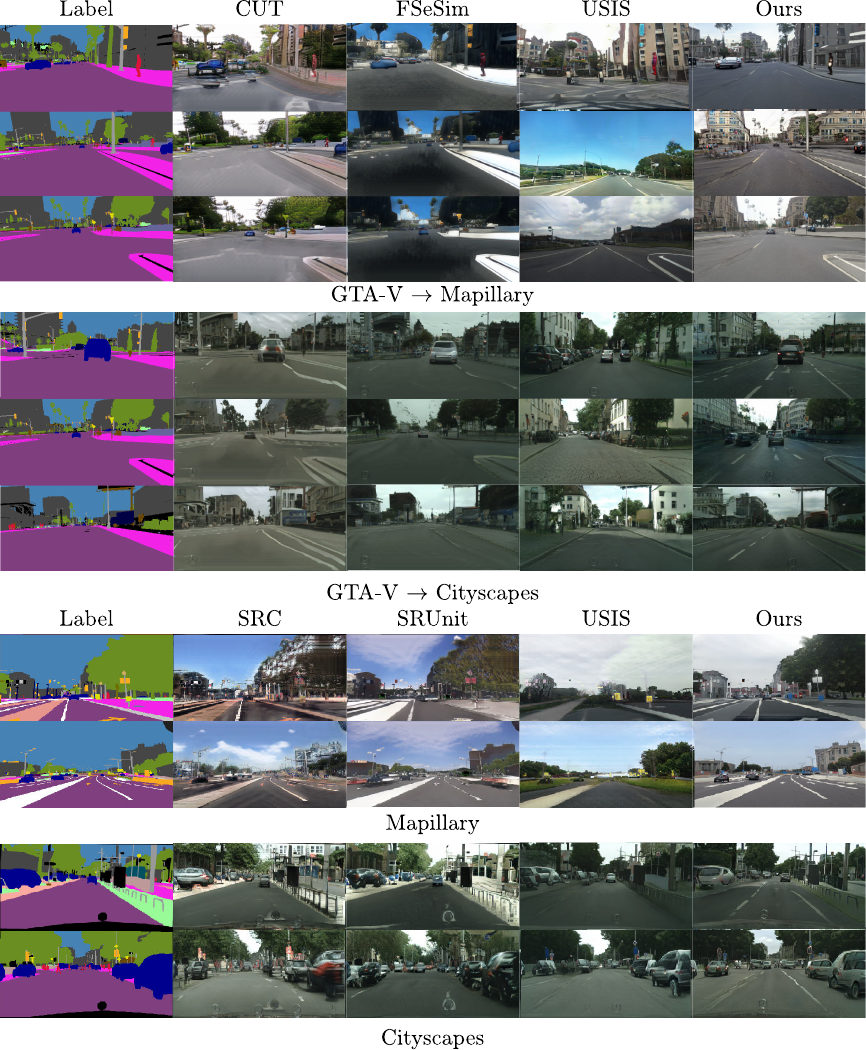}
\caption{ We show qualitative results for our model and other baselines on GTA-V $\rightarrow$ Mapillary, GTA-V $\rightarrow$ Cityscapes, Mapillary, Cityscapes }
\label{fig:qualitative}
\end{figure*}

We compare against state-of-the-art unpaired generative models~\cite{cut, sesim, srunit, src, usisicassp}. Our method is able to outperform existing methods in alignment in all settings by a large margin and is the best or second best in terms of image quality (FID, KID). We also notice that the proposed approach generalizes well when the input is a label from the real dataset (see Results on Cityscapes and Mapillary Val. splits). We show qualitative results on Cityscapes and Mapillary in Figure~\ref{fig:qualitative}.
\subsection{Ablation Studies}
\textbf{How to preserve alignment with the input label?} In Table~\ref{table:abl_alignment}, we compare between different alignment strategies. In Config A, we perform LPIPS on the whole image instead of patches; in Config B and C, LPIPS is performed on $\mathcal{P}(x)$ with 4 and 16 patches, respectively. All patches in the same experiment have equal size, i.e., in experiment B, patches have an area of $25\%$ of the total image's area. In Config D, we try a different approach: instead of generating whole images and aligning individual patches, we generate and align individual patches. In this setting, the training is done only on the patch level, which takes a substantially longer time to train. Results show that the hybrid approach of generating images and aligning patches is more optimal than generating and aligning images only or patches only. We find that the patch size is also a very important design parameter; very small patches do not provide enough context for alignment. 
\begin{table}[h!]
\setlength{\tabcolsep}{0.2em}
\renewcommand{\arraystretch}{0.95}
\centering

\begin{tabular}{|c|c|c|ccc|}
    \hline
    Config & $\mathcal{D}_u$ & $\{\mathcal{D}_l\}_{l=1}^L$ & FID $\downarrow$  &  mIoU$\uparrow$ & KID $\downarrow$  \\
    \hline 
     A & Patches (4) & Patches (4) & 60.7 & 24.6 & 0.026 \\
     B & Patches (4) & Whole Image & 72.8 & 20.6 & 0.030 \\
     C & Whole Image & Whole Image & 51.8 & 21.4 & 0.018 \\
     D & Whole Image & Patches (16) & 57.8 & 26.3 & 0.023 \\
     E & Whole Image & Patches (4) & \textbf{40.4} & \textbf{27.1} & \textbf{0.012} \\ 
     
     \hline
    \end{tabular}
    \caption{Comparison between different image discrimination strategies.}
\label{table:abl_disc}
    \vspace{-1em}
\end{table}
\begin{table}[h!]
	\setlength{\tabcolsep}{0.2em}
	\renewcommand{\arraystretch}{0.95}
	\centering
	
	\adjustbox{width=1.0\linewidth}{\begin{tabular}{|c|ccc|cc|}
        \hline
         \multirow{2}{*}{Method} & \multicolumn{3}{c|}{GTA-V $\rightarrow$ Cityscapes} & \multicolumn{2}{c|}{Cityscapes Val. Split} \\
         & FID $\downarrow$  &  mIoU$\uparrow$ & KID $\downarrow$ & FID $\downarrow$  &  mIoU$\uparrow$   \\
        \hline 
        Proposed Method   & \textbf{40.4} & 27.1 & \textbf{0.012} & \textbf{67.3} & 40.6 \\ 
        Two-Stage method  & 81.1 & \textbf{40.4} & 0.044 & 113.8 & \textbf{45.2} \\ 
        \hline
	\end{tabular}}
	\caption{How to use the synthetic image as an intermediary for         generating a photorealistic image? Comparison of our method with 
         a simple alternative.}
	\label{table:abl_guide}
        \vspace{-1em}
\end{table}

 \begin{table*}[t!]
\setlength{\tabcolsep}{0.2em}
\renewcommand{\arraystretch}{0.95}
\centering

\begin{tabular}{c|c|cc|cc}
    \multirow{2}{*}{Method } & \multirow{2}{*}{Dataset(s) } & \multicolumn{2}{c}{Mapillary Val. Split} & \multicolumn{2}{c}{Cityscapes Val. Split}    \\
    & &  FID $\downarrow$  &  mIoU$\uparrow$ &  FID $\downarrow$  &  mIoU$\uparrow$  \\
    
    \hline 

    \multirow{2}{*}{CUT~\cite{cut}} & Single Domain & 49.8 & 20.3 & 57.3 & 29.8 \\
    & Cross Domain & 73.4 & 16.5 & 75.4 & 29.6 \\
    
    \hline

    \multirow{2}{*}{SRUnit~\cite{srunit}} & Single Domain & 54.8 & 16.7 & 74.8 & 21.8 \\
    & Cross Domain & 70.8 & 16.5 & 82.1 & 23.4 \\
    
    \hline

    \multirow{2}{*}{F-SeSim~\cite{sesim}} & Single Domain & 74.1 &21.1 & 76.9 & 23.0 \\
    & Cross Domain & 92.9 & 15.5 & 111.1 & 24.4 \\
    
    \hline

    \multirow{2}{*}{SRC~\cite{src}} & Single Domain & 97.0 &17.6 & 64.2 & 31.2 \\
    & Cross Domain & 106.6 & 10.4 & 78.6 & 31.2 \\
    
    \hline

    \multirow{2}{*}{USIS~\cite{usisicassp}} & Single Domain & \best{23.8} &\best{ 31.0} & \best{53.7} & \best{44.8}  \\
    & Cross Domain & 54.6 & 14.8 & \second{54.5} & 28.3 \\

    \hline
    Ours & Cross Domain & \second{44.2} & \second{24.0} & 67.3 & \second{40.6} \\

\end{tabular}
\caption{Performance of different methods when trained in ideal conditions vs. when trained in a true unpaired setting. We show the strong performance of our model, trained on Cross Domain. The best results are marked in red; the second best are marked in blue. }
\label{table:study}
    \vspace{-1em}
\end{table*}



\textbf{Which discrimination strategy to use? } In Table~\ref{table:abl_disc}, we compare different choices for image discrimination. In Config A, we present whole images to the wavelet-based discriminator $\mathcal{D}_u$ and to the discriminator ensemble $\{\mathcal{D}_l\}_{l=1}^L$. In Config B, we present only patches as input to both discriminators, while in Config C and D, we experiment with hybrid approaches. Interestingly, whole image discrimination only (Config A) leads to a smaller FID and KID than Config B and C, but it demonstrates the poorest alignment. Config C and D achieve high alignment scores, but again a larger patch size proves to be essential for better diversity and alignment. 

\textbf{How to use the synthetic image as an intermediary for generating a photorealistic image? } A key idea of our approach is to use the synthetic image as a guide to the content in the generated image because it can provide helpful spatial information not present in the input label map (edges and corners, boundaries between different objects, shadows, texture). We have used the synthetic image in the alignment loss through a perceptual loss function. However, a very straightforward method to exploit the label map, is to learn the mapping between synthetic labels and images in a supervised way, then learn the mapping between the generated synthetic images and real images, using unpaired I2I model. In this "two-stage" approach, we employ OASIS~\cite{oasis} as the supervised SIS model and VSAIT~\cite{vsait} as the unpaired Synthetic-to-Real I2I model. Results in Table~\ref{table:abl_guide} reveal that while the alignment of this simple two-stage approach is better than our model, the FID is very high. Moreover, it doesn't generalize well to real labels, as the FID remains very high, and the difference in mIoU is reduced, which justifies our approach.

 \textbf{On the utility of the proposed approach and task} We compare the performance of our model against the state-of-the-art trained on the Synthetic-to-Real dataset and on the Real dataset only in an unpaired manner. We report the performance on the validation split of Mapillary. Results show that the performance of previous works substantially drops when their training and test distributions differ. However, the proposed approach achieves high image fidelity and strong alignment on the real labels in test-time, although they were unseen during training. More interestingly, the performance of our model surpasses 4 state-of-the-art models trained on the original dataset and comes only second to USIS. This implies that a carefully designed generative model trained on synthetic data can generalize well to real data.  

\section{Conclusion}
\label{sec:conclusion}
In this work, we explore a new task that exploits synthetic data to train a generative model for SIS, substantially alleviating labeling costs without sacrificing photorealism. Compared to synthetic-to-real I2I, this task has the advantage of allowing easier manipulation in the image space post-generation. We presented a framework that outperforms previous works on this task and has interestingly shown a strong generalization ability when the test-time input labels are drawn from a different distribution than the labels seen during training. Most importantly, we have shown that a hybrid approach of discriminating both images and patches is key to bypassing the semantic domain gap between images and labels. Additionally, using the synthetic image as a guide to a patch content loss promotes stronger alignment without undermining the photorealism of generated images. We believe that the proposed task can offer a pragmatic setting for training generative models and encourage future works to explore how to use synthetic data to train generative models for stronger generalization.

\clearpage
\clearpage
\begingroup
\setstretch{0.9}
\setlength\bibitemsep{0pt}
\printbibliography

@inproceedings{spade,
	title={Semantic image synthesis with spatially-adaptive normalization},
	author={Park, Taesung and Liu, Ming-Yu and Wang, Ting-Chun and Zhu, Jun-Yan},
	booktitle={Conference on Computer Vision and Pattern Recognition (CVPR)},
	year={2019}
}

@inproceedings{ccfpse,
	title={Learning to predict layout-to-image conditional convolutions for semantic image synthesis},
	author={Liu, Xihui and Yin, Guojun and Shao, Jing and Wang, Xiaogang and others},
	booktitle={Advances in Neural Information Processing Systems (NeurIPS)},
	year={2019}
}

@InProceedings{gtav,
author = {Stephan R. Richter and Vibhav Vineet and Stefan Roth and Vladlen Koltun},
title = {Playing for Data: {G}round Truth from Computer Games},
booktitle = {European Conference on Computer Vision (ECCV)},
year = {2016},
editor = {Bastian Leibe and Jiri Matas and Nicu Sebe and Max Welling},
series = {LNCS}, 
volume = {9906}, 
publisher = {Springer International Publishing},
pages = {102--118}
}

@inproceedings{pix2pixhd,
	title={High-resolution image synthesis and semantic manipulation with conditional {GANs}},
	author={Wang, Ting-Chun and Liu, Ming-Yu and Zhu, Jun-Yan and Tao, Andrew and Kautz, Jan and Catanzaro, Bryan},
	booktitle={Conference on Computer Vision and Pattern Recognition (CVPR)},
	year={2018}
}

@inproceedings{cordts2016cityscapes,
	title={The cityscapes dataset for semantic urban scene understanding},
	author={Cordts, Marius and Omran, Mohamed and Ramos, Sebastian and Rehfeld, Timo and Enzweiler, Markus and Benenson, Rodrigo and Franke, Uwe and Roth, Stefan and Schiele, Bernt},
	booktitle={Conference on Computer Vision and Pattern Recognition (CVPR)},
	year={2016}
}

@inproceedings{yu2017dilated,
	title={Dilated residual networks},
	author={Yu, Fisher and Koltun, Vladlen and Funkhouser, Thomas},
	booktitle={Conference on Computer Vision and Pattern Recognition (CVPR)},
	year={2017}
}

@inproceedings{pix2pix,
  title={Image-to-image translation with conditional adversarial networks},
  author={Isola, Phillip and Zhu, Jun-Yan and Zhou, Tinghui and Efros, Alexei A},
	booktitle={Conference on Computer Vision and Pattern Recognition (CVPR)},
  year={2017}
}

@inproceedings{Ronneberger2015UNetCN,
  title={U-Net: Convolutional Networks for Biomedical Image Segmentation},
  author={O. Ronneberger and P. Fischer and T. Brox},
  booktitle={MICCAI},
  year={2015}
}

@article{sean,
  title={SEAN: Image Synthesis With Semantic Region-Adaptive Normalization},
  author={Peihao Zhu and Rameen Abdal and Yipeng Qin and Peter Wonka},
  journal={2020 IEEE/CVF Conference on Computer Vision and Pattern Recognition (CVPR)},
  year={2020},
  pages={5103-5112}
}

@article{clade,
  title={Rethinking Spatially-Adaptive Normalization},
  author={Tan, Zhentao and Chen, Dongdong and Chu, Qi and Chai, Menglei and Liao, Jing and He, Mingming and Yuan, Lu and Yu, Nenghai},
  journal={arXiv:2004.02867},
  year={2020}
}

@inproceedings{munit,
  title={Multimodal unsupervised image-to-image translation},
  author={Huang, Xun and Liu, Ming-Yu and Belongie, Serge and Kautz, Jan},
  booktitle={European Conference on Computer Vision (ECCV)},
  year={2018}
}

@INPROCEEDINGS{Alexey2016,
	title = {Generating Images with Perceptual Similarity Metrics based on Deep Networks},
	author = {Dosovitskiy, Alexey and Brox, Thomas},
	booktitle = {Advances in Neural Information Processing Systems (NeurIPs)},
	editor = {D. D. Lee and M. Sugiyama and U. V. Luxburg and I. Guyon and R. Garnett},
	year = {2016}
}

@INPROCEEDINGS{Gatys2016,	
	author={L. A. {Gatys} and A. S. {Ecker} and M. {Bethge}},	
	booktitle={Conference on Computer Vision and Pattern Recognition (CVPR)}, 	
	title={Image Style Transfer Using Convolutional Neural Networks}, 	
	year={2016}}

@inproceedings{palette,
  title={Semantic Palette: Guiding Scene Generation with Class Proportions},
  author={Le Moing, Guillaume and Vu, Tuan-Hung and Jain, Himalaya and P{\'e}rez, Patrick and Cord, Matthieu},
  booktitle={Proceedings of the IEEE/CVF Conference on Computer Vision and Pattern Recognition},
  pages={9342--9350},
  year={2021}
}

@inproceedings{carla,
  title={CARLA: An open urban driving simulator},
  author={Dosovitskiy, Alexey and Ros, German and Codevilla, Felipe and Lopez, Antonio and Koltun, Vladlen},
  booktitle={Conference on robot learning},
  pages={1--16},
  year={2017},
  organization={PMLR}
}

@inproceedings{cyclegan,
	title={Unpaired image-to-image translation using cycle-consistent adversarial networks},
	author={Zhu, Jun-Yan and Park, Taesung and Isola, Phillip and Efros, Alexei A},
	booktitle= {International Conference on Computer Vision (ICCV)},
	year={2017}
}

@inproceedings{drit,
	title={Diverse Image-to-Image Translation via Disentangled Representation},
	author={Lee, Hsin-Ying and Tseng, Hung-Yu and Huang, Jia-Bin and Singh, Maneesh Kumar and Yang, Ming-Hsuan},
	booktitle={European Conference on Computer Vision (ECCV)},
	year={2018}
}

@inproceedings{gcgan,
  title={Geometry-consistent generative adversarial networks for one-sided unsupervised domain mapping},
  author={Fu, Huan and Gong, Mingming and Wang, Chaohui and Batmanghelich, Kayhan and Zhang, Kun and Tao, Dacheng},
  booktitle={Conference on Computer Vision and Pattern Recognition (CVPR)},
  year={2019}
}

@inproceedings{distancegan,
  title={One-sided unsupervised domain mapping},
  author={Benaim, Sagie and Wolf, Lior},
  booktitle={Advances in Neural Information Processing Systems (NeurIPS)},
  year={2017}
}

@inproceedings{travelgan,
  title={Travelgan: Image-to-image translation by transformation vector learning},
  author={Amodio, Matthew and Krishnaswamy, Smita},
  booktitle={Proceedings of the IEEE Conference on Computer Vision and Pattern Recognition},
  pages={8983--8992},
  year={2019}
}

@article{stylegan,
  title={A Style-Based Generator Architecture for Generative Adversarial Networks},
  author={Tero Karras and S. Laine and Timo Aila},
  journal={2019 IEEE/CVF Conference on Computer Vision and Pattern Recognition (CVPR)},
  year={2019},
  pages={4396-4405}
}

@article{Gal2021SWAGANAS,
  title={SWAGAN: A Style-based Wavelet-driven Generative Model},
  author={Rinon Gal and Dana Cohen and Amit H. Bermano and D. Cohen-Or},
  journal={ArXiv},
  year={2021},
  volume={abs/2102.06108}
}

@inproceedings{
oasis,
title={You Only Need Adversarial Supervision for Semantic Image Synthesis},
author={Edgar Sch{\"o}nfeld and Vadim Sushko and Dan Zhang and Juergen Gall and Bernt Schiele and Anna Khoreva},
booktitle={International Conference on Learning Representations},
year={2021}
}

@inproceedings{cut,
  title={Contrastive Learning for Unpaired Image-to-Image Translation},
  author={Taesung Park and Alexei A. Efros and Richard Zhang and Jun-Yan Zhu},
  booktitle={European Conference on Computer Vision},
  year={2020}
}

@article{epe,
  title={Enhancing photorealism enhancement},
  author={Richter, Stephan R and AlHaija, Hassan Abu and Koltun, Vladlen},
  journal={IEEE Transactions on Pattern Analysis \& Machine Intelligence},
  volume={45},
  number={02},
  pages={1700--1715},
  year={2023},
  publisher={IEEE Computer Society}
}

@article{usiscag,
title = {USIS: Unsupervised Semantic Image Synthesis},
journal = {Computers \& Graphics},
year = {2023},
issn = {0097-8493},
doi = {https://doi.org/10.1016/j.cag.2022.12.010},
url = {https://www.sciencedirect.com/science/article/pii/S0097849323000018},
author = {George Eskandar and Mohamed Abdelsamad and Karim Armanious and Bin Yang},
}

@INPROCEEDINGS{usisicassp,
  author={Eskandar, George and Abdelsamad, Mohamed and Armanious, Karim and Zhang, Shuai and Yang, Bin},
  booktitle={ICASSP 2022 - 2022 IEEE International Conference on Acoustics, Speech and Signal Processing (ICASSP)}, 
  title={Wavelet-Based Unsupervised Label-to-Image Translation}, 
  year={2022},
  volume={},
  number={},
  pages={1760-1764},
  doi={10.1109/ICASSP43922.2022.9746759}}

@article{explainability,
  title={Explainability of deep vision-based autonomous driving systems: Review and challenges},
  author={Zablocki, {\'E}loi and Ben-Younes, H{\'e}di and P{\'e}rez, Patrick and Cord, Matthieu},
  journal={International Journal of Computer Vision},
  pages={1--28},
  year={2022},
  publisher={Springer}
}

@inproceedings{srunit,
  title={Semantically robust unpaired image translation for data with unmatched semantics statistics},
  author={Jia, Zhiwei and Yuan, Bodi and Wang, Kangkang and Wu, Hong and Clifford, David and Yuan, Zhiqiang and Su, Hao},
  booktitle={Proceedings of the IEEE/CVF International Conference on Computer Vision},
  pages={14273--14283},
  year={2021}
}

@inproceedings{sesim,
  title={The spatially-correlative loss for various image translation tasks},
  author={Zheng, Chuanxia and Cham, Tat-Jen and Cai, Jianfei},
  booktitle={Proceedings of the IEEE/CVF Conference on Computer Vision and Pattern Recognition},
  pages={16407--16417},
  year={2021}
}

@inproceedings{src,
  title={Exploring Patch-wise Semantic Relation for Contrastive Learning in Image-to-Image Translation Tasks},
  author={Jung, Chanyong and Kwon, Gihyun and Ye, Jong Chul},
  booktitle={Proceedings of the IEEE/CVF Conference on Computer Vision and Pattern Recognition},
  pages={18260--18269},
  year={2022}
}

@inproceedings{mapillary,
  title={The mapillary vistas dataset for semantic understanding of street scenes},
  author={Neuhold, Gerhard and Ollmann, Tobias and Rota Bulo, Samuel and Kontschieder, Peter},
  booktitle={Proceedings of the IEEE international conference on computer vision},
  pages={4990--4999},
  year={2017}
}

@article{sbgan,
  title={Semantic bottleneck scene generation},
  author={Azadi, Samaneh and Tschannen, Michael and Tzeng, Eric and Gelly, Sylvain and Darrell, Trevor and Lucic, Mario},
  journal={arXiv preprint arXiv:1911.11357},
  year={2019}
}

@article{edgesis,
  title={Edge guided GANs with semantic preserving for semantic image synthesis},
  author={Tang, Hao and Qi, Xiaojuan and Xu, Dan and Torr, Philip HS and Sebe, Nicu},
  journal={arXiv preprint arXiv:2003.13898},
  year={2020}
}

@inproceedings{decomposing,
  title={Decomposing image generation into layout prediction and conditional synthesis},
  author={Volokitin, Anna and Konukoglu, Ender and Van Gool, Luc},
  booktitle={Proceedings of the IEEE/CVF Conference on Computer Vision and Pattern Recognition Workshops},
  pages={372--373},
  year={2020}
}

@article{projected,
  title={Projected gans converge faster},
  author={Sauer, Axel and Chitta, Kashyap and M{\"u}ller, Jens and Geiger, Andreas},
  journal={Advances in Neural Information Processing Systems},
  volume={34},
  pages={17480--17492},
  year={2021}
}

@article{vgg,
  title={Very deep convolutional networks for large-scale image recognition},
  author={Simonyan, Karen and Zisserman, Andrew},
  journal={arXiv preprint arXiv:1409.1556},
  year={2014}
}

@article{ganinpainting,
  title={Globally and locally consistent image completion},
  author={Iizuka, Satoshi and Simo-Serra, Edgar and Ishikawa, Hiroshi},
  journal={ACM Transactions on Graphics (ToG)},
  volume={36},
  number={4},
  pages={1--14},
  year={2017},
  publisher={ACM New York, NY, USA}
}

@inproceedings{vsait,
  title={Unpaired Image Translation via Vector Symbolic Architectures},
  author={Theiss, Justin and Leverett, Jay and Kim, Daeil and Prakash, Aayush},
  booktitle={Computer Vision--ECCV 2022: 17th European Conference, Tel Aviv, Israel, October 23--27, 2022, Proceedings, Part XXI},
  pages={17--32},
  year={2022},
  organization={Springer}
}
\endgroup

\end{document}